\documentclass[conference]{IEEEtran}
\IEEEoverridecommandlockouts
\usepackage{cite}
\usepackage{amsmath,amssymb,amsfonts}
\usepackage{algorithmic}
\usepackage{graphicx}
\usepackage{textcomp}
\usepackage{xcolor}
\def\BibTeX{{\rm B\kern-.05em{\sc i\kern-.025em b}\kern-.08em
    T\kern-.1667em\lower.7ex\hbox{E}\kern-.125emX}}
\begin{document}

\title{HuBo-VLM: Unified Vision-Language Model designed for HUman roBOt interaction tasks\\
{\footnotesize \textsuperscript{*}Note: Technical Report}
}

\author{\IEEEauthorblockN{1\textsuperscript{st} Zichao Dong}
\IEEEauthorblockA{\textit{UDeer.ai} \\
zichao@udeer.ai}
\and
\IEEEauthorblockN{2\textsuperscript{nd} Weikun Zhang}
\IEEEauthorblockA{\textit{Zhejiang University} \\
zhangwk@zju.edu.cn}
\and
\IEEEauthorblockN{3\textsuperscript{rd} Xufeng Huang}
\IEEEauthorblockA{\textit{UDeer.ai} \\
xufeng@udeer.ai}
\and
\IEEEauthorblockN{4\textsuperscript{th} Hang Ji}
\IEEEauthorblockA{\textit{UDeer.ai} \\
jihang@udeer.ai}
\and
\IEEEauthorblockN{5\textsuperscript{th} Xin Zhan}
\IEEEauthorblockA{\textit{UDeer.ai} \\
zhanxin@udeer.ai}
\and
\IEEEauthorblockN{6\textsuperscript{th} Junbo Chen *}
\IEEEauthorblockA{\textit{UDeer.ai} \\
junbo@udeer.ai}

\thanks{This is a draft version contains Talk2Car benchmark only.}

}

\maketitle

\begin{abstract}
Human robot interaction is an exciting task, which aimed to guide robots following instructions from human. Since huge gap lies between human natural language and machine codes, end to end human robot interaction models is fair challenging. Further, visual information receiving from sensors of robot is also a hard language for robot to perceive. In this work, HuBo-VLM is proposed to tackle perception tasks associated with human robot interaction including object detection and visual grounding by a unified transformer based vision language model. Extensive experiments on the Talk2Car benchmark demonstrate the effectiveness of our approach. Code would be publicly available in https://github.com/dzcgaara/HuBo-VLM. 
\end{abstract}


\section{Introduction}
Human machine interaction tasks are important challenging tasks since it is hard for machine to understand intentions of human. However, there is no doubt that machine would become more intelligence and helpful if it could end to end taking actions with sensor data and human instruction as input. Thanks to previous successful works major in multi-modality tasks, it is possible for one single model to let robot take human natural language as instructions and images from raw sensor as input while output response corresponding to instructions.   

Nevertheless, there still drawbacks in previous VLMs. As for models in VLBert style, region of interests of images are required to be extracted in advance, which could constrain these methods becoming end to end. As for vision-centric models like MDETR, it is hard for it to unify diverse type of tasks like image caption. In addition, it is hard for vision-centric models to acquire the ability of reasoning. Speaking of previous non-ROI language-centric VLMs like OFA\cite{wang2022ofa}, there is no instruction encoding mechanism which would lead difficulty in operate different tasks with same model parameters.   

To solve above problems, a novel HuBo-VLM architecture is proposed. It use a language model as the brain to reason and solve diverse tasks in human robot problems. An instruction encoder is designed to encode human instruction, which could provide wide scalability facing multiple tasks. 

\section{RELATED WORK}

\subsection{BERT}
BERT\cite{devlin2018bert} is a deeply bidirectional Transformer\cite{vaswani2017attention} encoder pretrained on large amounts of unlabeled text. BERT uses two novel unsupervised pretraining tasks: masked language modeling, which predicts randomly masked words based on context, and next sentence prediction, which predicts if two sentences follow each other.  Unlike previous methods like ELMo\cite{peters2018deep} and OpenAI GPT\cite{radford2018improving} which use shallowly bidirectional LSTMs or unidirectional Transformers during pretraining, BERT's deep bidirectional Transformers allow each word to indirectly "see itself" in all layers, leading to stronger contextual representations. Through comprehensive experiments, Devlin et al.demonstrate the importance of BERT's bidirectional pretraining and find that BERTLARGE significantly outperforms BERTBASE across various tasks, especially those with little training data. They also show BERT's effectiveness in both fine-tuning and feature-based approaches.

\subsection{VL-BERT}
In the pursuit of aligning visual and linguistic clues for various tasks, the work by Su et al. introduces Visual-Linguistic BERT (VL-BERT)\cite{su2019vl}, a novel pre-trainable generic representation tailored for visual-linguistic tasks. Building upon the Transformer architecture, VL-BERT ingeniously extends its capabilities to accept both visual and linguistic embedded features. This integration allows for a more harmonized alignment between visual and linguistic elements, significantly enhancing the performance in downstream applications such as visual commonsense reasoning, visual question answering, and referring expression comprehension. Notably, VL-BERT's effectiveness is empirically validated as it secured the top position as a single model on the VCR\cite{zellers2019recognition} benchmark. The approach delineated in this work provides valuable insights into the synergistic fusion of visual and textual information, contributing to the broader landscape of multimodal learning.

\subsection{VilBERT}
In the realm of multimodal learning, the integration of visual and textual information has been a subject of extensive research. The work by Lu et al. presents ViLBERT (Vision-and-Language BERT)\cite{lu2019vilbert}, a pioneering model that learns task-agnostic representations of images and natural language. Extending the BERT\cite{su2019vl} architecture, ViLBERT processes visual and textual inputs separately and employs co-attentional transformer layers for their integration. Pretrained on the Conceptual Captions\cite{sharma2018conceptual} dataset, the model is then transferred to various vision-and-language tasks. This approach offers a substantial contribution to the understanding of how visual and linguistic information can be synergistically combined, setting a new benchmark in the field of multimodal learning.

\subsection{OFA}
Yuan et al. proposed OFA\cite{wang2022ofa}, a unified sequence-to-sequence framework for multimodal pretraining that is task-agnostic, modality-agnostic, and supports task comprehensiveness. OFA formulates both pretraining and finetuning tasks with handcrafted instructions, requiring no task-specific components. It is pretrained on 20M image-text pairs to support a diverse set of uni-modal and cross-modal tasks including generation, grounding, classification, and language modeling. A transformer acts as the shared compute engine for different modalities and tasks. Without using extra labeled data, OFA\cite{wang2022ofa} achieves state-of-the-art results on cross-modal downstream tasks like image captioning and VQA at that time, and is competitive with specialized models on uni-modal tasks. OFA\cite{wang2022ofa} also shows strong capability for zero-shot learning and domain adaptation without finetuning. The unified sequence-to-sequence formulation and lack of task-specific customization demonstrates OFA's potential as a generalist model applicable to both seen and unseen tasks across modalities. The ability to pretrain a capable visiolinguistic model on a relatively small dataset bodes well for the scalability of this approach.
\subsection{LLama}
The recent surge in the development of Large Language Models (LLMs) has led to remarkable advancements in the field of natural language processing. A noteworthy contribution in this context is the LLaMA\cite{touvron2023llama} series, a collection of foundation language models with a parameter range from 7B to 65B. Trained on publicly available datasets, LLaMA models demonstrate the feasibility of achieving state-of-the-art performance at that time without relying on proprietary or inaccessible datasets. For example, LLaMA-13B outperforms GPT-3 on various benchmarks, despite being significantly smaller in size. The LLaMA series also emphasizes training efficiency, focusing not only on the computational budget for training but also on the critical aspect of inference budget. This holistic approach ensures that the models are not only powerful but also efficient in real-world applications. The open-sourcing of the LLaMA models further contributes to the democratization of access and study of LLMs, making them a significant milestone in the ongoing evolution of language modeling.
\subsection{DETR}

The visual detection task involves predicting regions of interest and their corresponding categories within an image. 
Based on Convolutional Neural Networks (CNNs), methods have indeed achieved remarkable success in the field of object detection. However, these approaches often involve a significant amount of manually designed components, such as anchor boxes, sliding windows, and non-maximum suppression (NMS)\cite{ren2015faster,tian2019fcos}.
Recently, Transformer-based methods have been gradually replacing CNN-based methods in the field of object detection. This shift has been facilitated by approaches like Vision Transformer (ViT)\cite{dosovitskiy2020image}, Swim\cite{liu2021swin} and SwimV2\cite{liu2022swin}. These Transformer-based models leverage the self-attention mechanism to capture relationships between different image regions more effectively and have demonstrated promising results in various object detection tasks. 
DETR is the first model to incorporate a Transformer-based approach into the detection head\cite{sun2021makes}. During training, it associates learnable queries with each instance through Hungarian matching, setting queries without matches as unlabeled. In the inference phase, it decodes each query to compute scores, generating detection boxes and corresponding class labels by applying a threshold.

\subsection{MDETR}
Unlike CNN-based methods, Transformers can handle sequences of data, making them well-suited for processing both image and text data, which can be particularly advantageous in tasks involving multi-modal information.
In the realm of multi-modal object detection and reasoning, several approaches have been proposed to overcome the limitations of traditional systems. Some of these methods aim to associate text queries with images to better understand and process multi-modal data.
Past research has attempted to integrate object detection with natural language processing (NLP) techniques. These methods often employ separate models for image and text processing and then combine their results. However, this separation can lead to information loss and inconsistencies.
In MDETR\cite{kamath2021mdetr}, the authors employ a dataset consisting of 1.3 million image-text pairs for pre-training the model. This extensive dataset ensures that the model learns to extract consistent features from paired images and texts.With the success of Transformer models in NLP, some research has begun to explore the application of Transformer models to multi-modal tasks. MDETR\cite{kamath2021mdetr} is a prominent example of this trend, employing a Transformer architecture that allows the model to perform end-to-end joint reasoning between images and text, capturing better correlations between the two modalities.

\subsection{MaskFormer}
The MaskFormer\cite{cheng2021per} approach represents a paradigm shift in semantic and panoptic segmentation, addressing both instance- and semantic-level tasks with a single mask classification model. Contrary to the conventional per-pixel classification techniques, MaskFormer operates on the principle that mask classification is sufficiently versatile to encompass various segmentation tasks. This method predicts a set of binary masks, each corresponding to a global class label prediction, thus providing a unified and simple framework. The superiority of MaskFormer is particularly evident in large vocabulary datasets, where it has set new benchmarks such as 55.6 mIoU on ADE20K\cite{zhou2017scene} and 52.7 PQ on COCO\cite{lin2014microsoft}. By converting any existing per-pixel classification model into a mask classification, MaskFormer not only simplifies the segmentation landscape but also enhances efficiency.

\subsection{VisionLLM}
The emergence of VisionLLM\cite{wang2023visionllm} marks a significant milestone in the pursuit of a unified framework for vision and language tasks. By aligning vision-centric tasks with LLM methodologies, VisionLLM introduces a unique perspective that enables the customization of tasks through language instructions. This innovative approach overcomes the limitations imposed by traditional vision foundation models (VFMs), which often struggle with open-ended task capabilities. VisionLLM's core components include a language-guided image tokenizer and an LLM-based decoder that orchestrates various tasks using language instructions. The framework's flexibility extends to fine-grained object-level customization as well as coarse-grained task-level customization. With impressive results, such as a 60\% mAP score on the COCO dataset, VisionLLM sets a new baseline for generalist vision and language models, highlighting the potential for unified modeling in this interdisciplinary field.

\subsection{Instruct BLIP}
InstructBLIP\cite{dai2023instructblip} represents a pioneering approach in the field of vision-language models, focusing on instruction tuning to create a unified natural language interface capable of addressing diverse vision-language tasks. Unlike previous work that often relied on limited visual components or multitask learning, InstructBLIP conducts a methodical study on vision-language instruction tuning, transforming 26 datasets into instruction tuning format. The framework's key innovation lies in its instruction-aware Query Transformer, which extracts visual features in accordance with specific instructions, thereby enhancing the model's adaptability to various tasks. The success of InstructBLIP is evident in its substantial outperformance of existing models such as BLIP-2\cite{li2023blip2} and larger Flamingo\cite{alayrac2022flamingo} models, achieving state-of-the-art zero-shot performance and accuracy levels as high as 90.7\% on specific tasks like ScienceQA\cite{lu2022learn} questions with image contexts. The open-sourcing of InstructBLIP models further contributes to the ongoing discourse in the field, setting new benchmarks for vision-language tasks.

\section{METHOD}
\subsection{Overview}

\begin{figure*}[htbp]
    \centering
    \includegraphics[width=16cm]{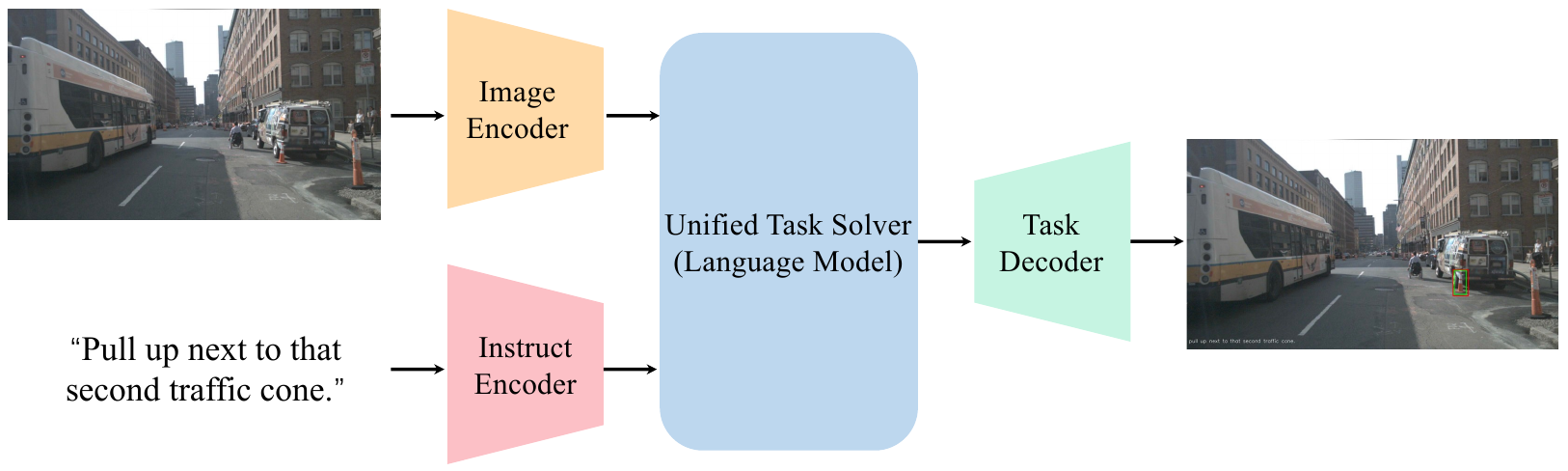}
    \caption{\small \textbf{Overall pipeline of HuBo-VLM}. Our model takes images and instructions as inputs. The images and instructions are separately processed by encoders before being fed into the Unified Task Solver. Subsequently, the outputs from the Task Decoder provide the final results. As shown here, we input an image along with the instruction "Pull up next to that second cone." The model ultimately generates a 2D bounding box that encompasses the corresponding cone. The green box represents the ground truth, while the red box indicates the model's output.}
    \label{fig:model}
\end{figure*}

Our proposed HuBo-VLM mainly contains three major components as shown in Fig.~\ref{fig:model}. Our model takes images and instuctions as input
A general image encoder is constructed in order to encode images as visual feature embedding. In parallel, human instructions as natural language sentences will be fed into our instruction encoder with a instruction embedding as output. Further, a unified task solver would take above two embedding as input. Our unified task solver is typical sequence to sequence style who is built by pure transformer architecture. Finally, a task related post processor is implemented in order to cooperate with our instruction encoder for diverse task decoding. It is worth noticing that training strategy is all of vital importance for our HuBo-VLM. Unified pre-training and task related instruct fine-tuning are main strategies to boost our model. Above mentioned parts would be depicted in following sections. 

\subsection{Image encoder}
Since we aimed in solving a vision language task, a image encoder is undoubtedly essential in our model. Similar to previous vision language models like BLIP and OFA\cite{wang2022ofa}, a general state of art image backbone without specify design for our task would satisfy our needs. As for IO of this component, the image encoder takes an image as input and outputs patches of image embedding. In our practice, transformer based methods like ViT and convolution neural network based methods like ResNet are both wise choices. Notice that we do not need extra ROI extraction like VL-BERT\cite{su2019vl}, which could helper us build an end to end unified human machine interaction model. By the way, for efficient fine-tuning, a frozen pre-trained image backbone could also served as a fair image encoder in our HuBo-VLM.

\subsection{Instruction encoder}
In order to equip our HuBo-VLM with open-set multi-task ability, an instruction encoder is proposed to encoder human instruction language as unified instruction embedding. For instance,  human would give a prompt like “Detect the person in blue hat by a bounding box in (x0, y0, x1, y1) style". The instruction embedding is an essential clue for combining basic image features from image encoder by latter attention mechanism in our following unified task solver. Inspired by Instruct-BLIP, instruction encoder part could also serve as a short-cut for efficient fine-tuning. We could consider image encoder as a task agnostic unified visual feature extractor while task related information could be decoupled by instruction embedding. We should notice that instruction encoder is extremely tiny compared with above image encoder. 

\subsection{Unified task solver}
Inspired by modern vision-language large language model manner, we suppose that language model is more suitable for serving as a brain in AGI system. Since there are tremendous amount of data for language model to learn ways to operate reasoning, visual model has limited amount of data while visual data also has more redundancy information. Considering efficiency for inference with limited robot embedded hardware, we did not use large language model like LLama for now. A typical encoder-decoder style transformer is utilized as our unified task solver while we could witness that it is strong enough to overwhelm all previous for visual grounding tasks in Talk2Car benchmark.  

It is also worth noticing that we define diverse human machine interaction tasks as a unified language sequence generation task. Take visual grounding task as an example, different as former state of art object detection based multi-modality method like MDETR\cite{kamath2021mdetr}, there is no traditional object detection design including regression loss and anchor mechanism. Thus, it is implement-friendly for expanding other custom open-set tasks without construction.  

\subsection{Task related post-processor}
After unified task solver, a sequence is generated corresponding to user instruction and visual clue. User would develop custom post process logic in order to transform the generated sequence to the original format that fit user needs. Take object detection as an example, user could convert the sentence “(x0, y0, x1, y1)" to a bounding box and then decorate it on the image.

\subsection{Unified pre-training}
Similar to previous multi-modality works like OFA\cite{wang2022ofa}, multiple existing multi-modality dataset in different tasks are firstly used as pre-training. Thanks to our unified task solver who has unified output format diverse task to unique sequence generation, we can use same architecture and weight to accomplish above diverse tasks. To be specific,  Conceptual Caption 12M (CC12M)\cite{91},  Conceptual Captions (CC3M)\cite{92}, SBU\cite{93} , MSCOCO image captions (COCO)\cite{74}, Visual Genome Captions (VG Captions)\cite{94}, VQAv2\cite{95}, GQA\cite{96}, RefCOCO\cite{75}, RefCOCO+\cite{75}, RefCOCOg\cite{76}, YFCC100M\cite{97}, ImageNet-21K\cite{81}, OpenImages\cite{98} ,Object365\cite{99} and Pile\cite{100} are mixed and used as pre-train dataset for our HuBo-VLM.

\subsection{Task related instruct finetuning}
After unified pre-training, custom annotated dataset is used to perform supervised finetuning. Note that the model is identical as pre-training stage. In our practice, around 1k data is fairly enough for a new task. Besides, we notice that a small learning rate would perform better if amount of custom data is small. We will take Talk2Car as an example and show our results on it in the next section.

\section{Experiment}

In this section we mainly introduce our experiments on Talk2Car visual grounding task. Extensive visualization results of our HuBo-VLM are shown in figure2. 

 \subsection{Dataset}
The Talk2Car dataset \cite{deruyttere2019talk2car} is the first dataset containing object reference instructions written in natural language for autonomous driving commands. Constructed based on the nuScenes dataset \cite{caesar2020nuscenes}, the Talk2Car dataset comprises 11,959 commands extracted from 850 videos within the nuScenes training set. Among these commands, 55.94\% pertain to videos captured in Boston, while 44.06\% relate to videos captured in Singapore. On average, each command consists of 11.01 words, 2.32 nouns, 2.29 verbs, and 0.62 adjectives. There are an average of 14.07 commands per video. The training, validation, and test sets encompass 8,349 (69.8\%), 1,163 (9.7\%), and 2,447 (20.4\%) samples, respectively \cite{deruyttere2019talk2car}.

\subsection{Implementation Details}
A vanilla  ResNet152 is used as our image encoder with input shape 800x1333. Bert is utilized as our instruct encoder and unified task solver. We use a small learning rate as 1e-5 during fine-tuning stage. End to end training is implemented while there are no parameter fronzon in our model. The training time is around 12hours with 8xV100.

\subsection{Quantitive Evaluation}

\begin{table}[htbp]
\centering
\caption{}
\label{table:leaderboard}
\begin{tabular}{ll}
\hline
\multicolumn{1}{c}{\textbf{Model}} & \multicolumn{1}{c}{\textbf{AP50}} \\ \hline
\textbf{HuBo-VLM}                  & \textbf{76.74}                    \\
Deformerable-MDETR                 & 74.4                              \\
Stacked VLBert                     & 71                                \\
CMRT                               & 69.1                              \\
Vilbert (Base)                     & 68.9                              \\
CMSVG                              & 68.6                              \\
ASSMR                              & 66                                \\
AttnGrounder                       & 63.3                              \\
VL-Bert (Base)                     & 63.1                              \\
MSRR                               & 60.04                             \\
MAC                                & 50.51                             \\
SCRC                               & 38.7                              \\
OSM                                & 35.31                             \\
STACK-NMN                          & 33.71                             \\ \hline
\end{tabular}
\end{table}

The quantitative experimental results of our model on the Talk2Car dataset are presented in Table~\ref{table:leaderboard}. The evaluation metric used is AP50. It is evident from the results that our model's performance is highly competitive.

\subsection{Visualization and discussion}

\begin{figure*}[htbp]
\centering     

\begin{minipage}{0.3\linewidth}
    \vspace{3pt}
    \centerline{\includegraphics[width=\textwidth]{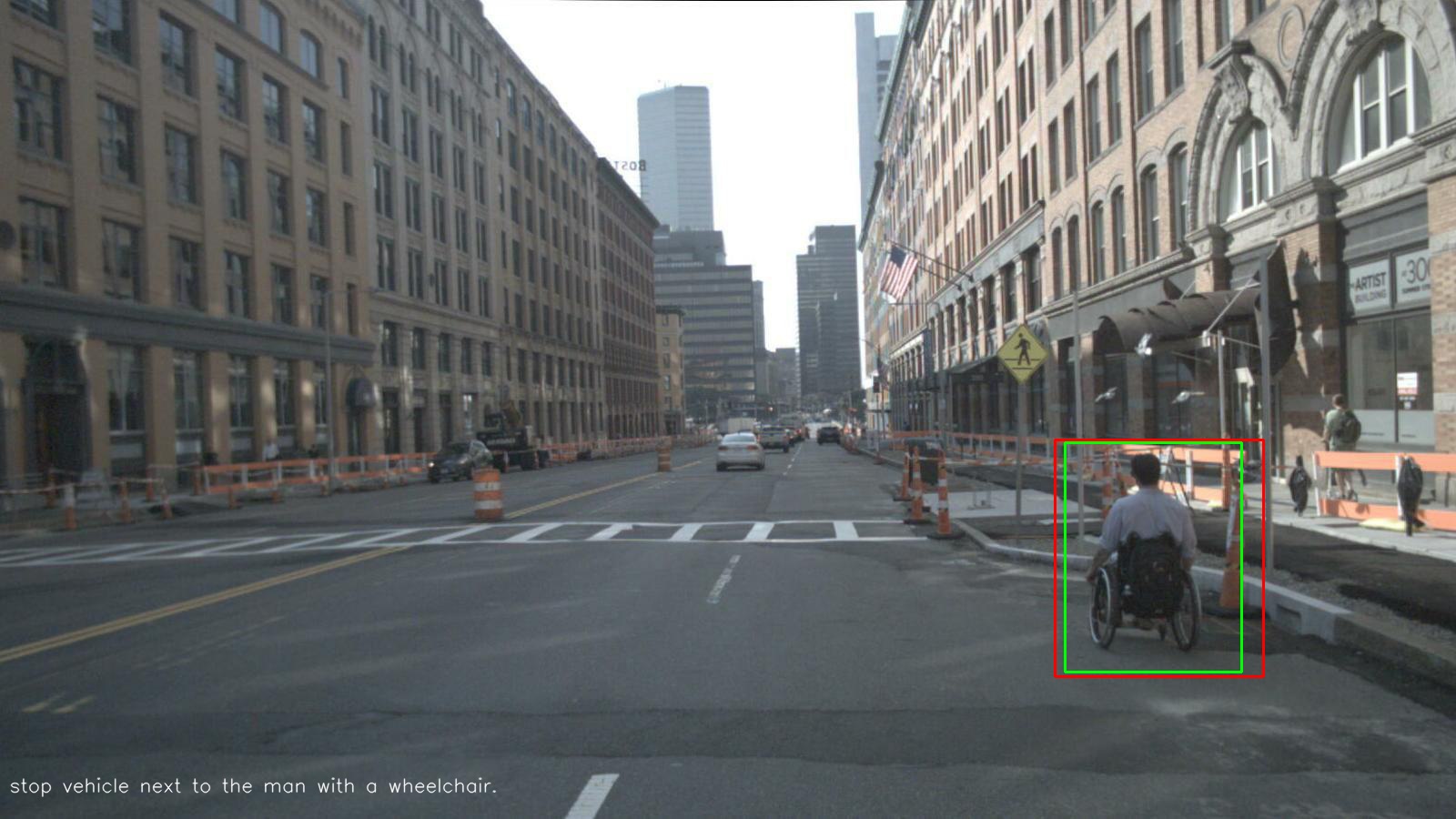}}
\end{minipage}
\begin{minipage}{0.3\linewidth}
    \vspace{3pt}
    \centerline{\includegraphics[width=\textwidth]{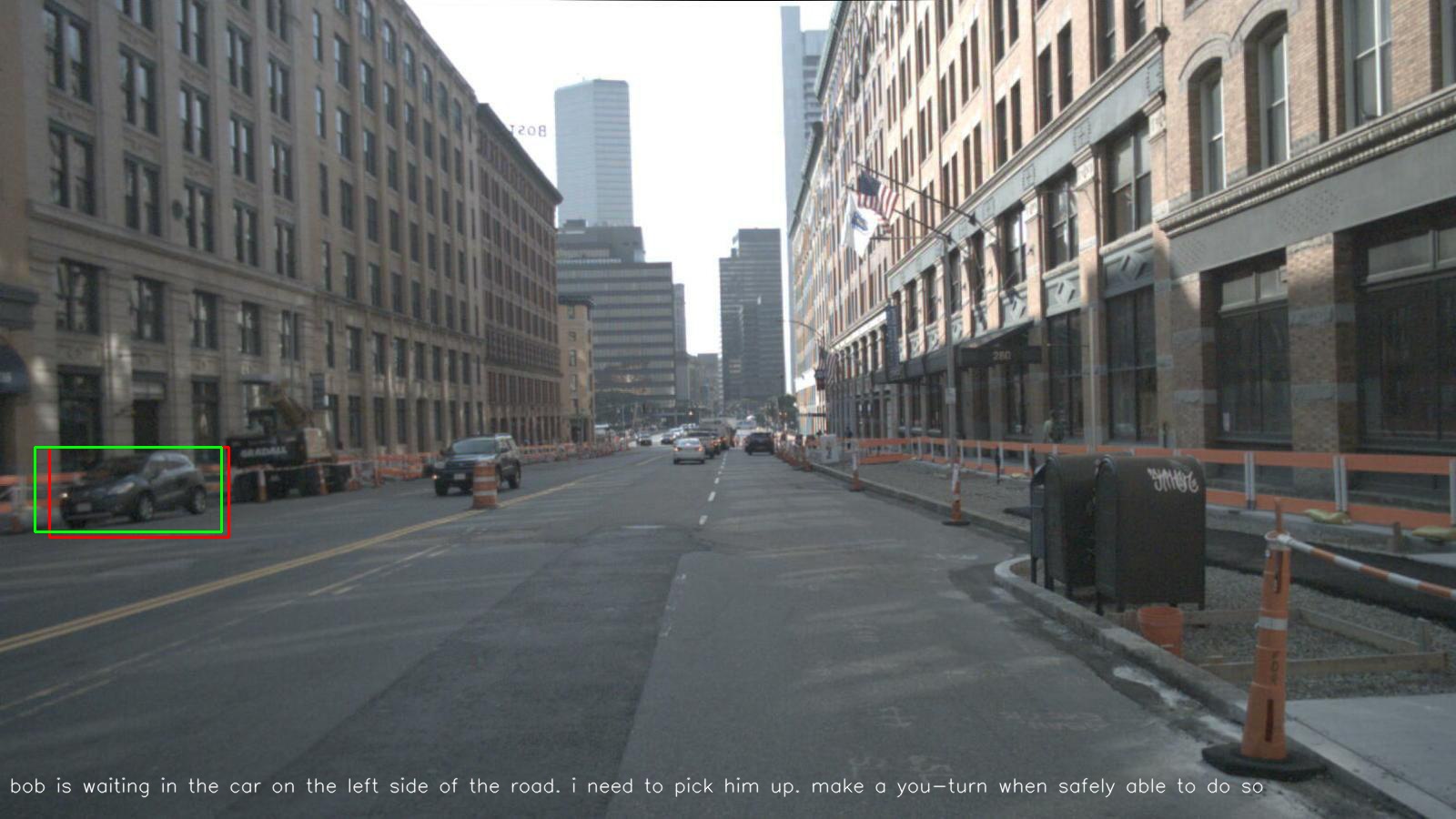}}
\end{minipage}
\begin{minipage}{0.3\linewidth}
    \vspace{3pt}
    \centerline{\includegraphics[width=\textwidth]{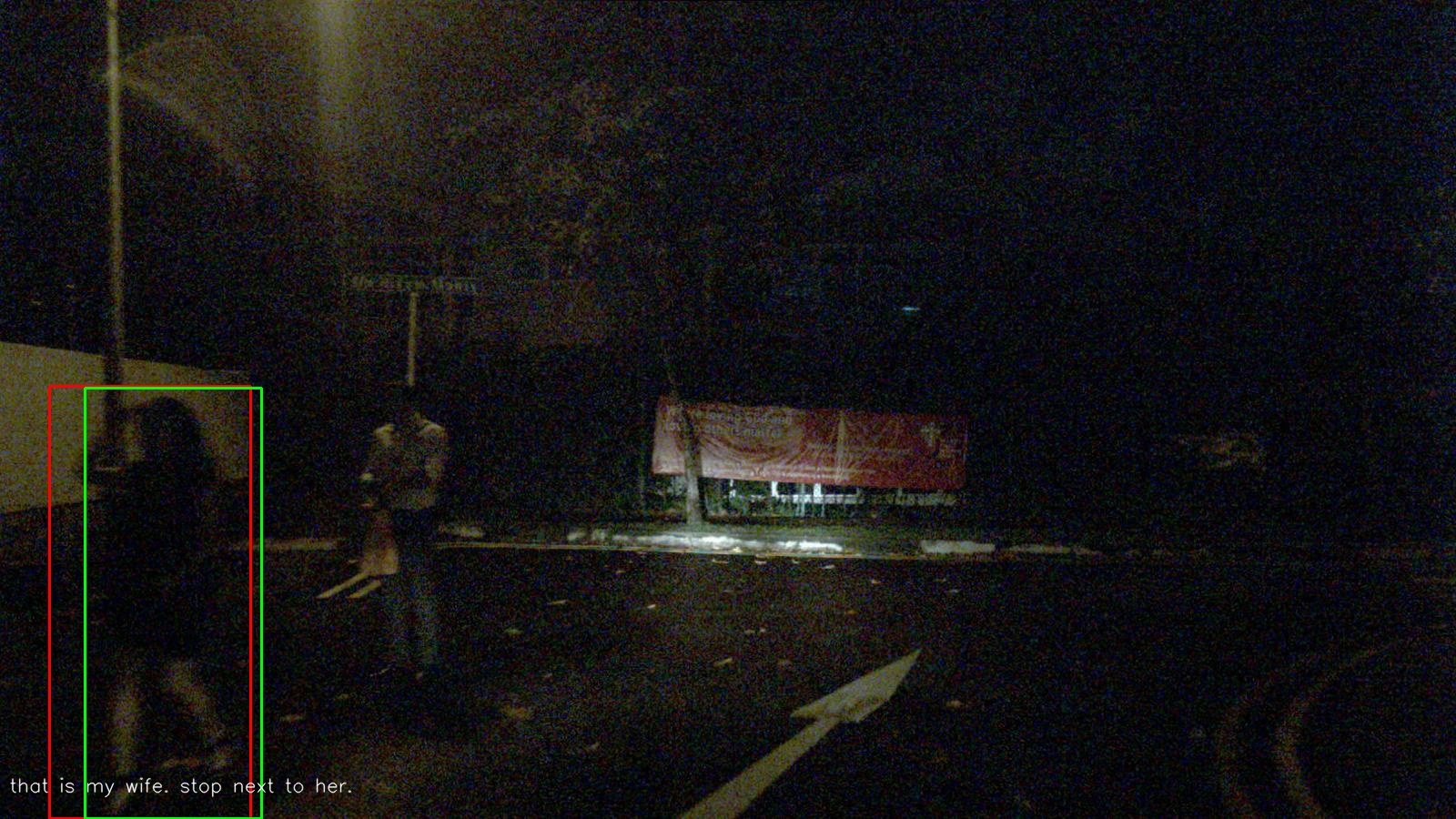}}
\end{minipage}

\begin{minipage}{0.3\linewidth}
\vspace{3pt}
    ``stop vehicle next to the man with a wheelchair."
\end{minipage}
\begin{minipage}{0.3\linewidth}
\vspace{3pt}
    ``bob is waiting in the car on the left side of the road. i need to pick him up. make a you-turn when safely able to do so."
\end{minipage}
\begin{minipage}{0.3\linewidth}
\vspace{3pt}
    ``that is my wife. stop next to her."
\end{minipage}

\begin{minipage}{0.3\linewidth}
    \vspace{3pt}
    \centerline{\includegraphics[width=\textwidth]{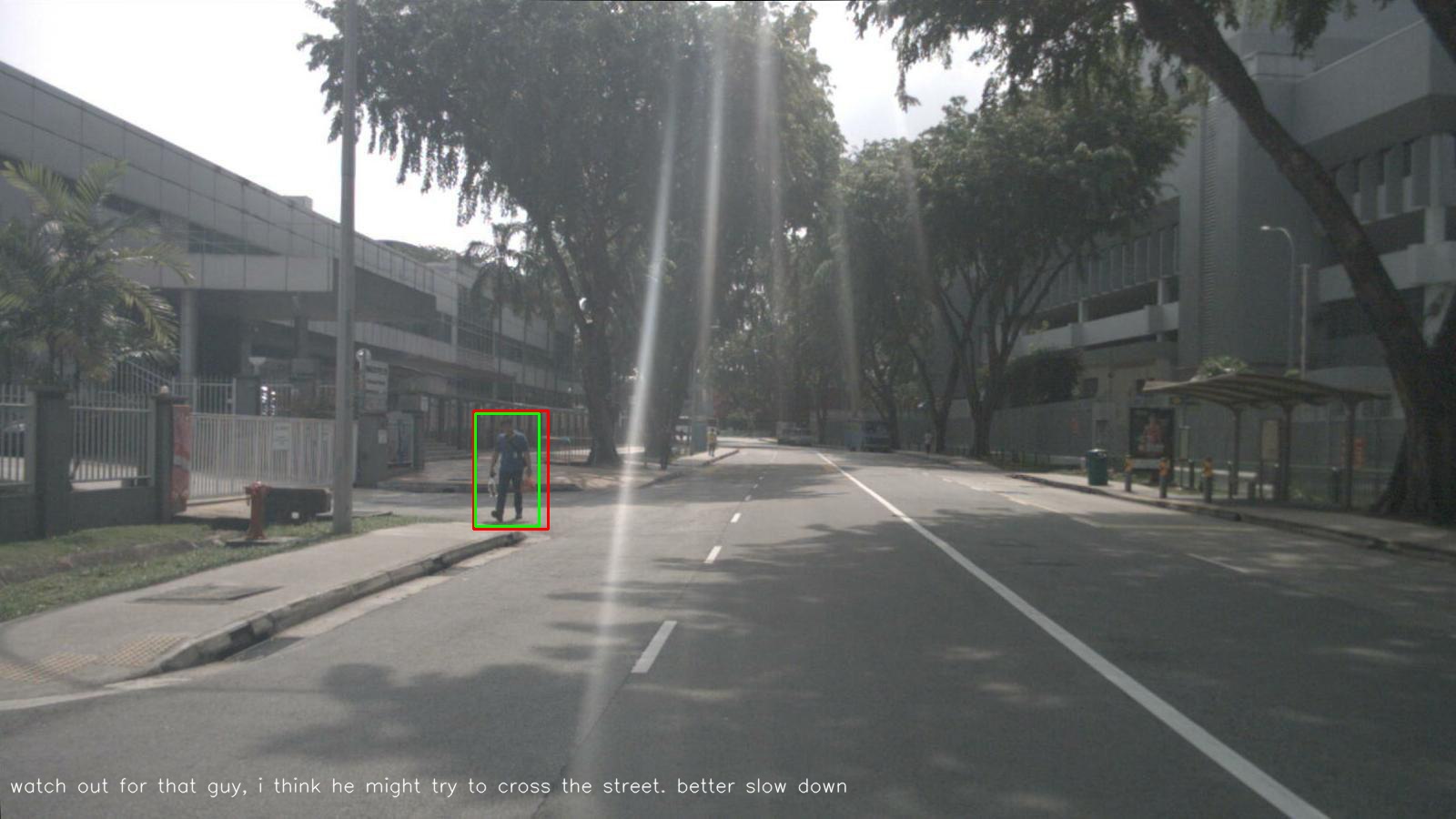}}
\end{minipage}
\begin{minipage}{0.3\linewidth}
    \vspace{3pt}
    \centerline{\includegraphics[width=\textwidth]{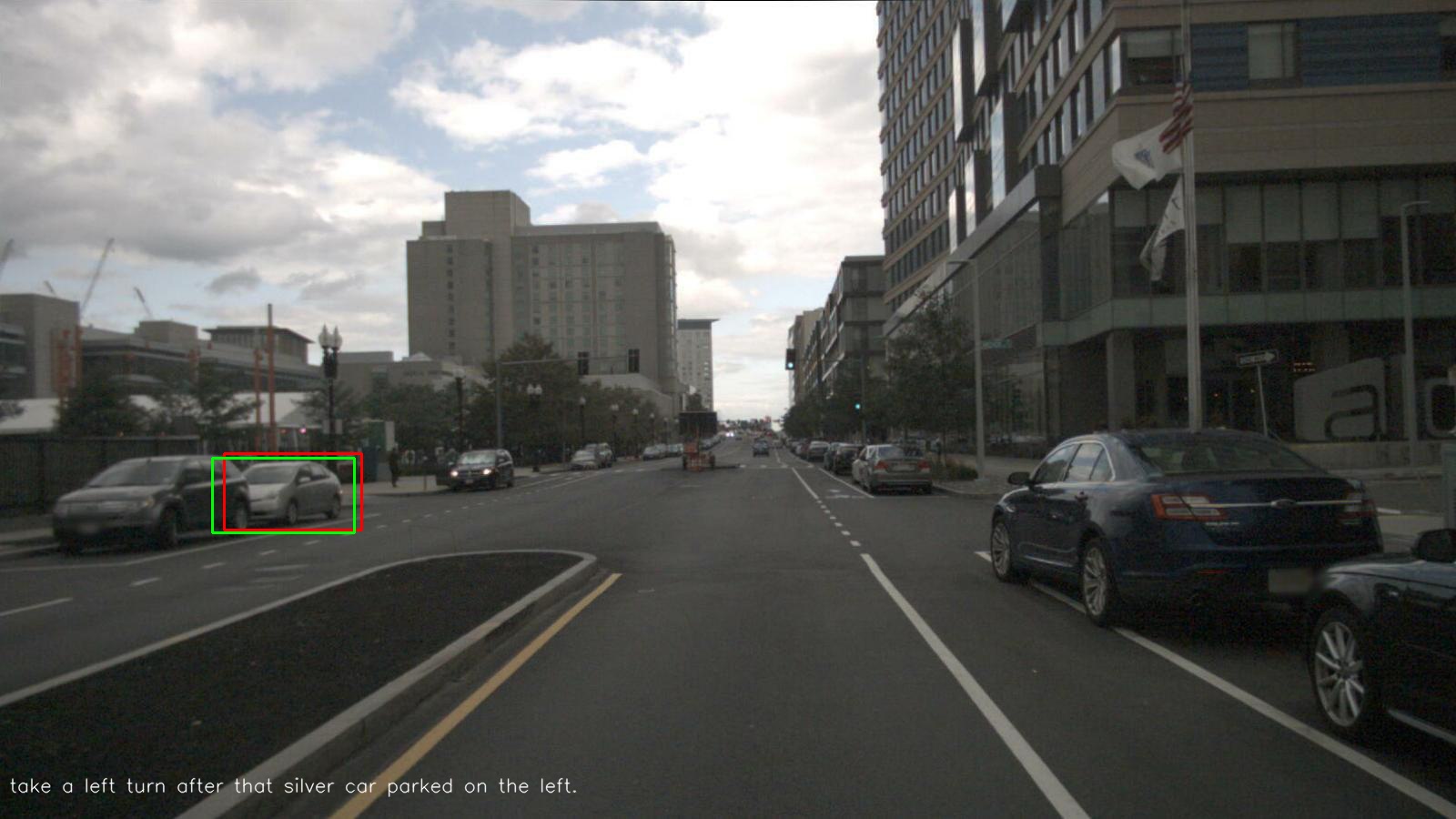}}
\end{minipage}
\begin{minipage}{0.3\linewidth}
    \vspace{3pt}
    \centerline{\includegraphics[width=\textwidth]{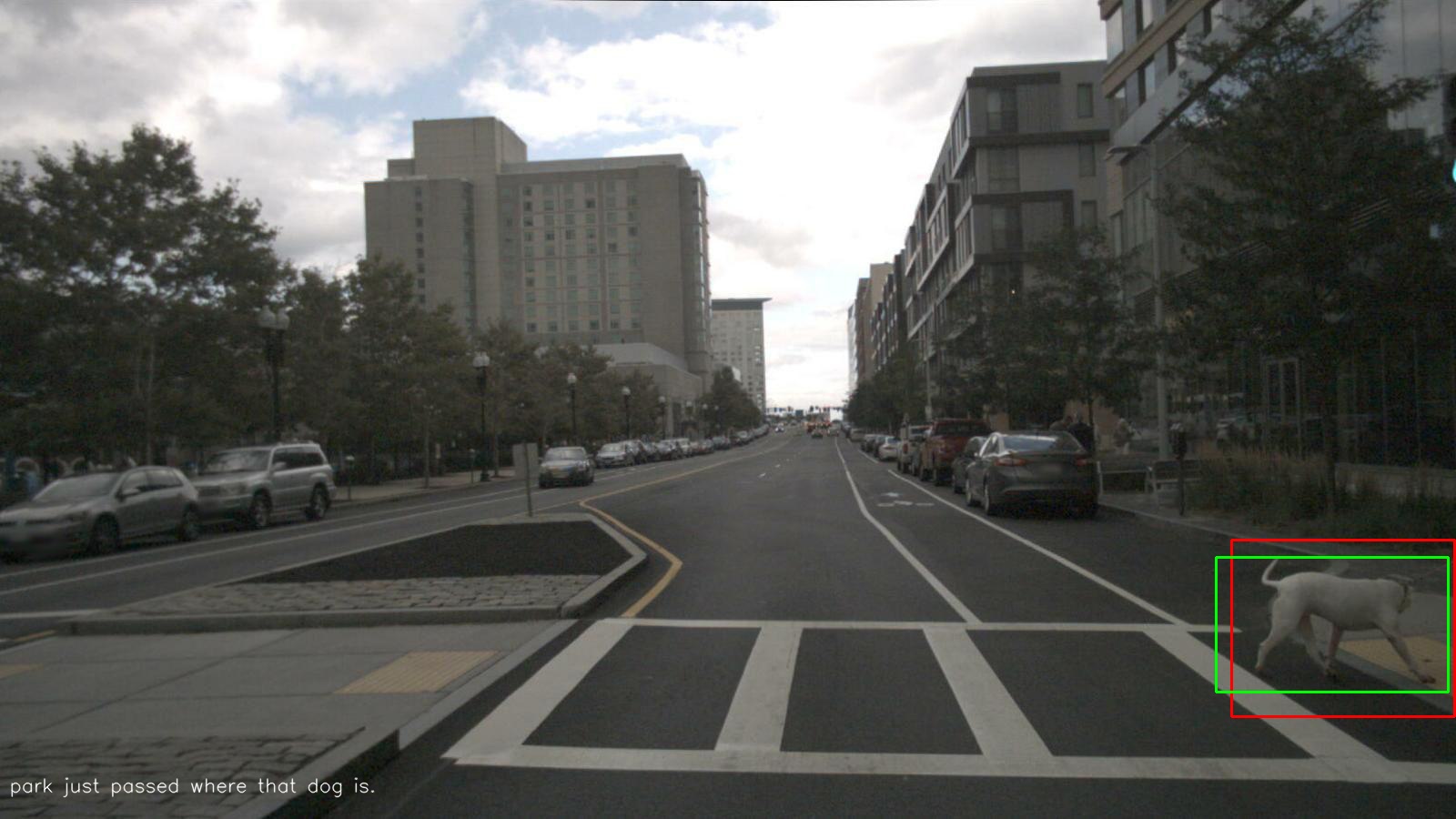}}
\end{minipage}

\begin{minipage}{0.3\linewidth}
\vspace{3pt}
    ``watch out for that guy, i think he might try to cross the street. better slow down."
\end{minipage}
\begin{minipage}{0.3\linewidth}
\vspace{3pt}
    ``take a left turn after that silver car parked on the left."
\end{minipage}
\begin{minipage}{0.3\linewidth}
\vspace{3pt}
    ``park just passed where that dog is."
\end{minipage}

\begin{minipage}{0.3\linewidth}
    \vspace{3pt}
    \centerline{\includegraphics[width=\textwidth]{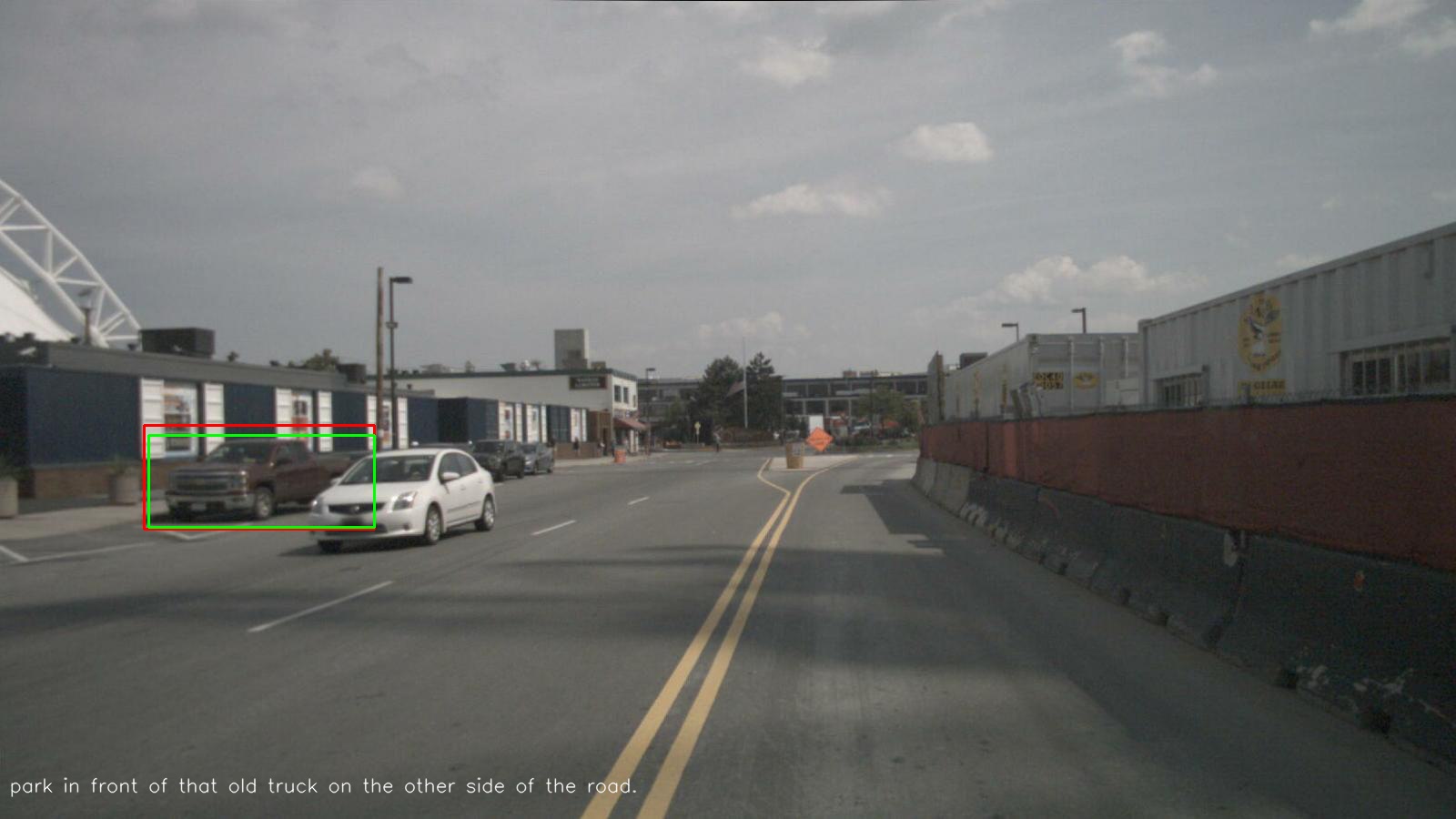}}
\end{minipage}
\begin{minipage}{0.3\linewidth}
    \vspace{3pt}
    \centerline{\includegraphics[width=\textwidth]{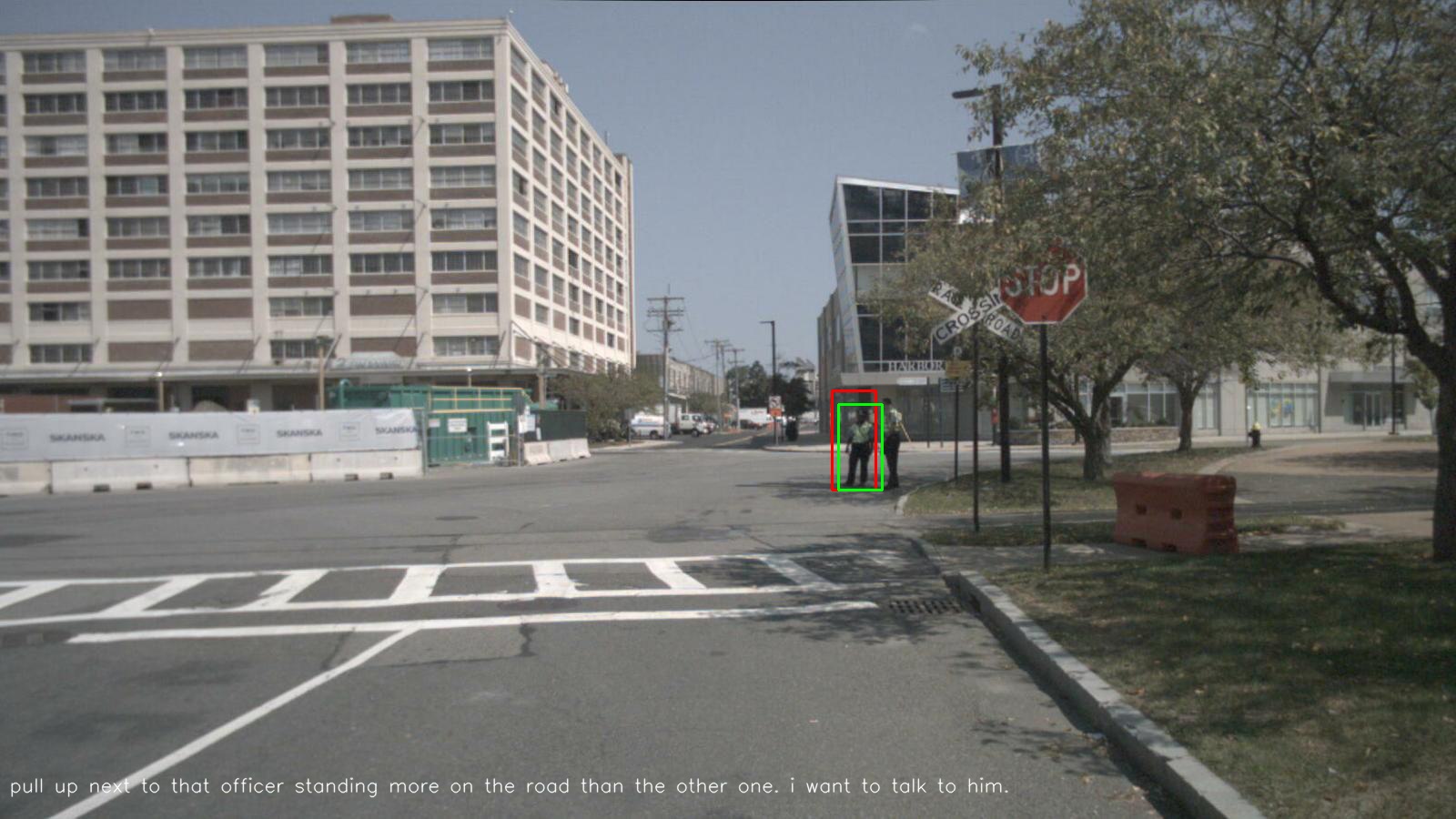}}
\end{minipage}
\begin{minipage}{0.3\linewidth}
    \vspace{3pt}
    \centerline{\includegraphics[width=\textwidth]{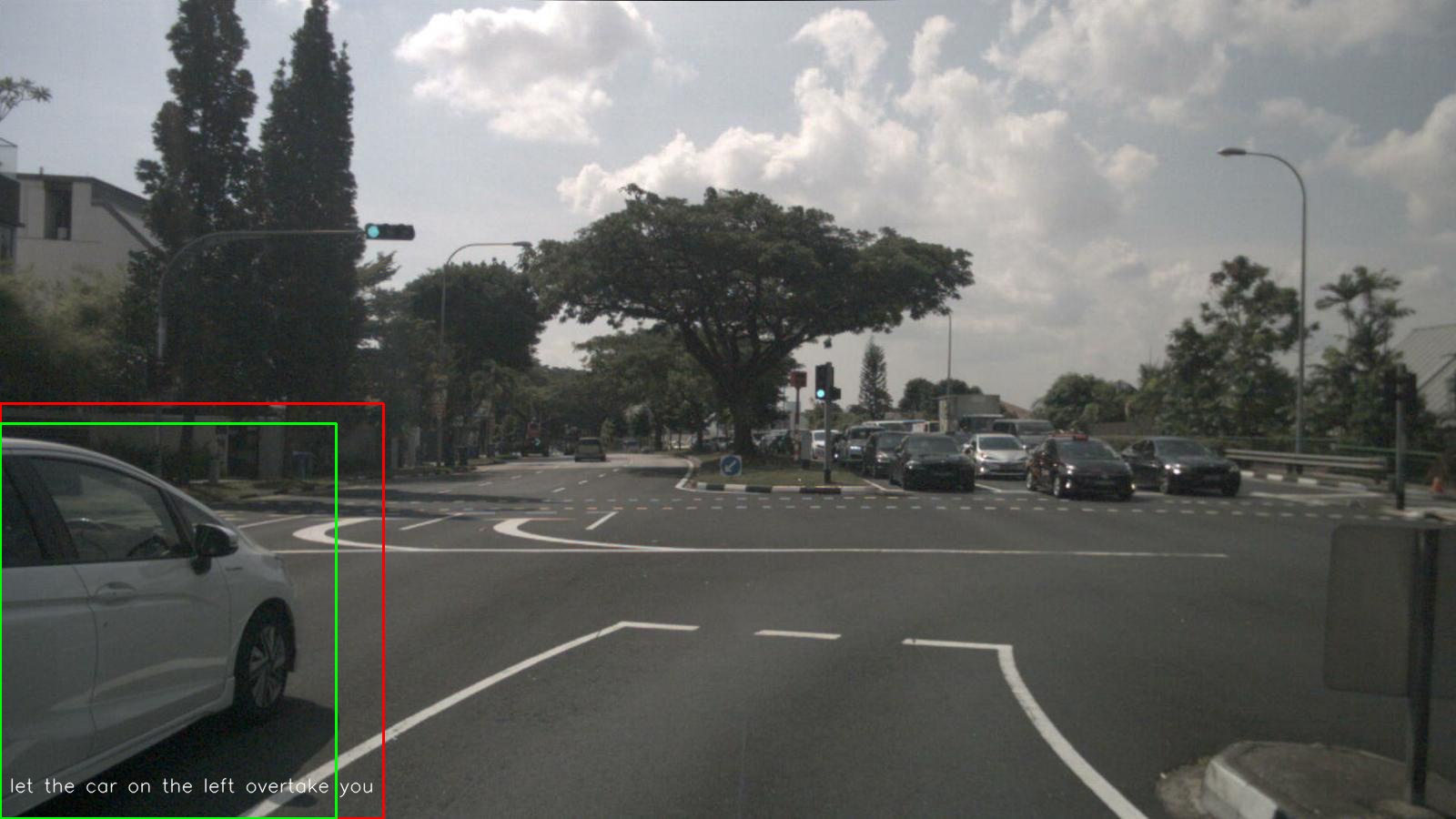}}
\end{minipage}

\begin{minipage}{0.3\linewidth}
\vspace{3pt}
    ``par in front of that old truck on the other side of the road."
\end{minipage}
\begin{minipage}{0.3\linewidth}
\vspace{3pt}
    ``pull up next to that officer standing more on the road than the other one. i want to talk to him."
\end{minipage}
\begin{minipage}{0.3\linewidth}
\vspace{3pt}
    ``let the car on the left overtake you."
\end{minipage}

\begin{minipage}{0.3\linewidth}
    \vspace{3pt}
    \centerline{\includegraphics[width=\textwidth]{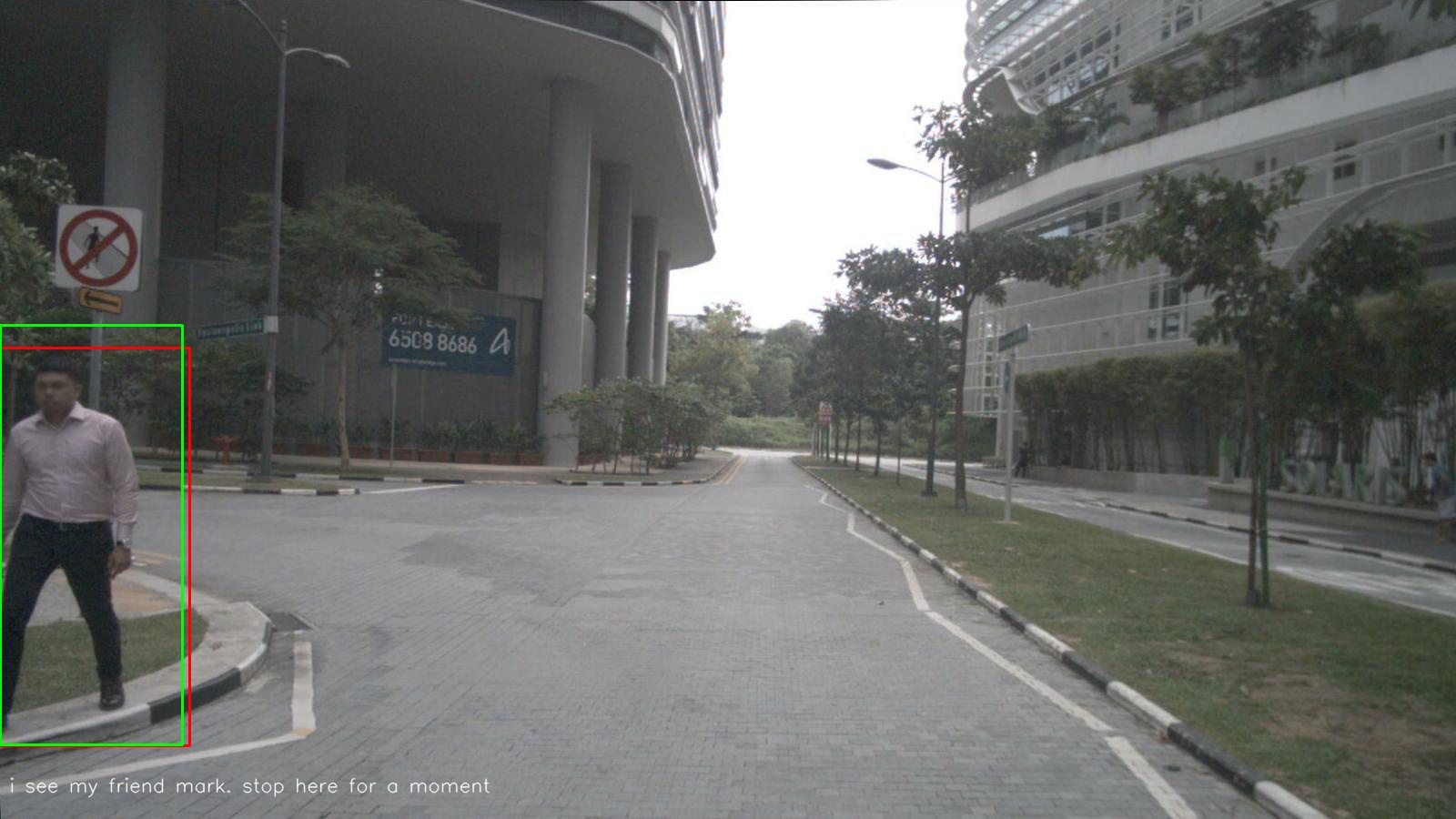}}
\end{minipage}
\begin{minipage}{0.3\linewidth}
    \vspace{3pt}
    \centerline{\includegraphics[width=\textwidth]{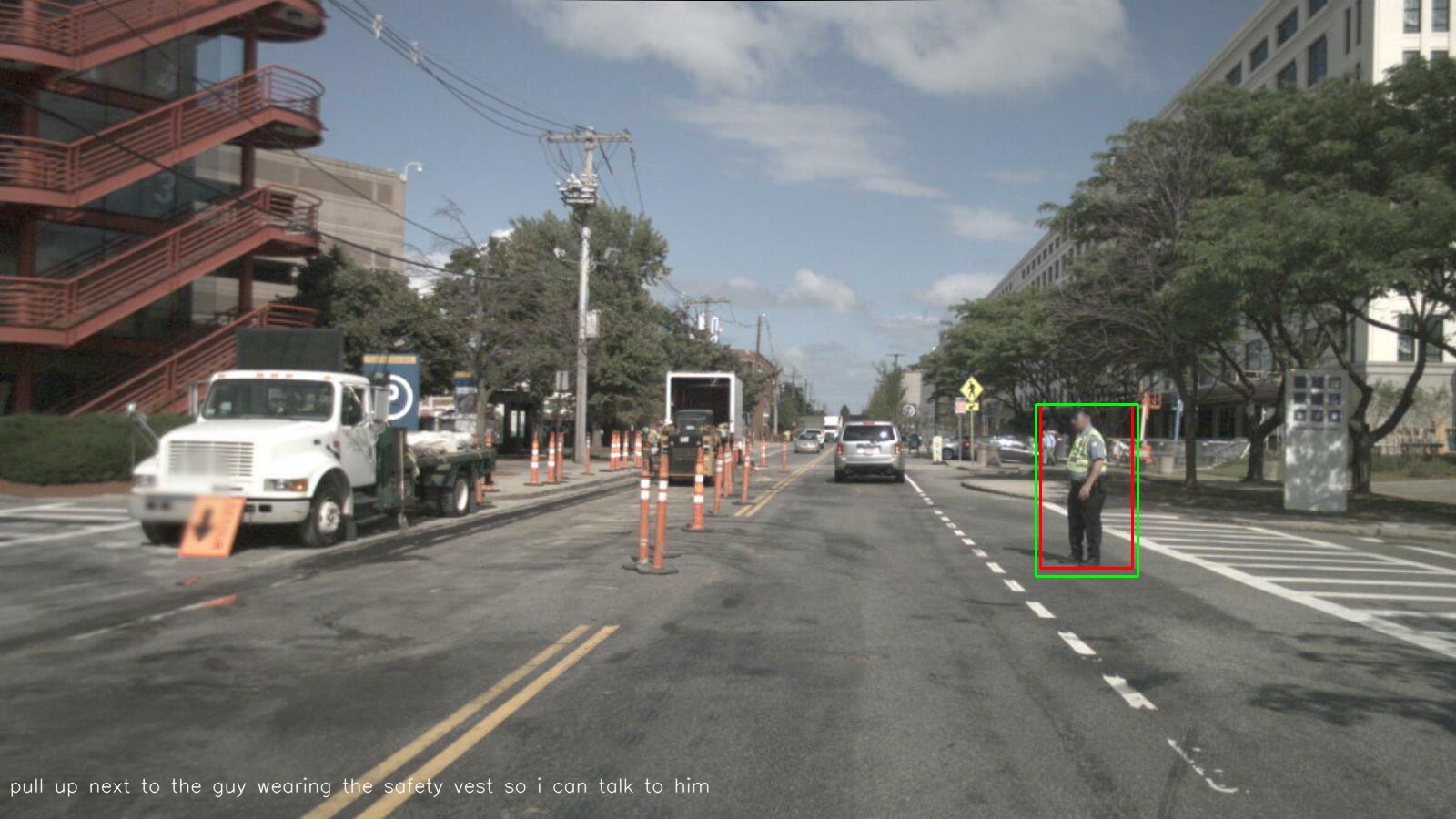}}
\end{minipage}
\begin{minipage}{0.3\linewidth}
    \vspace{3pt}
    \centerline{\includegraphics[width=\textwidth]{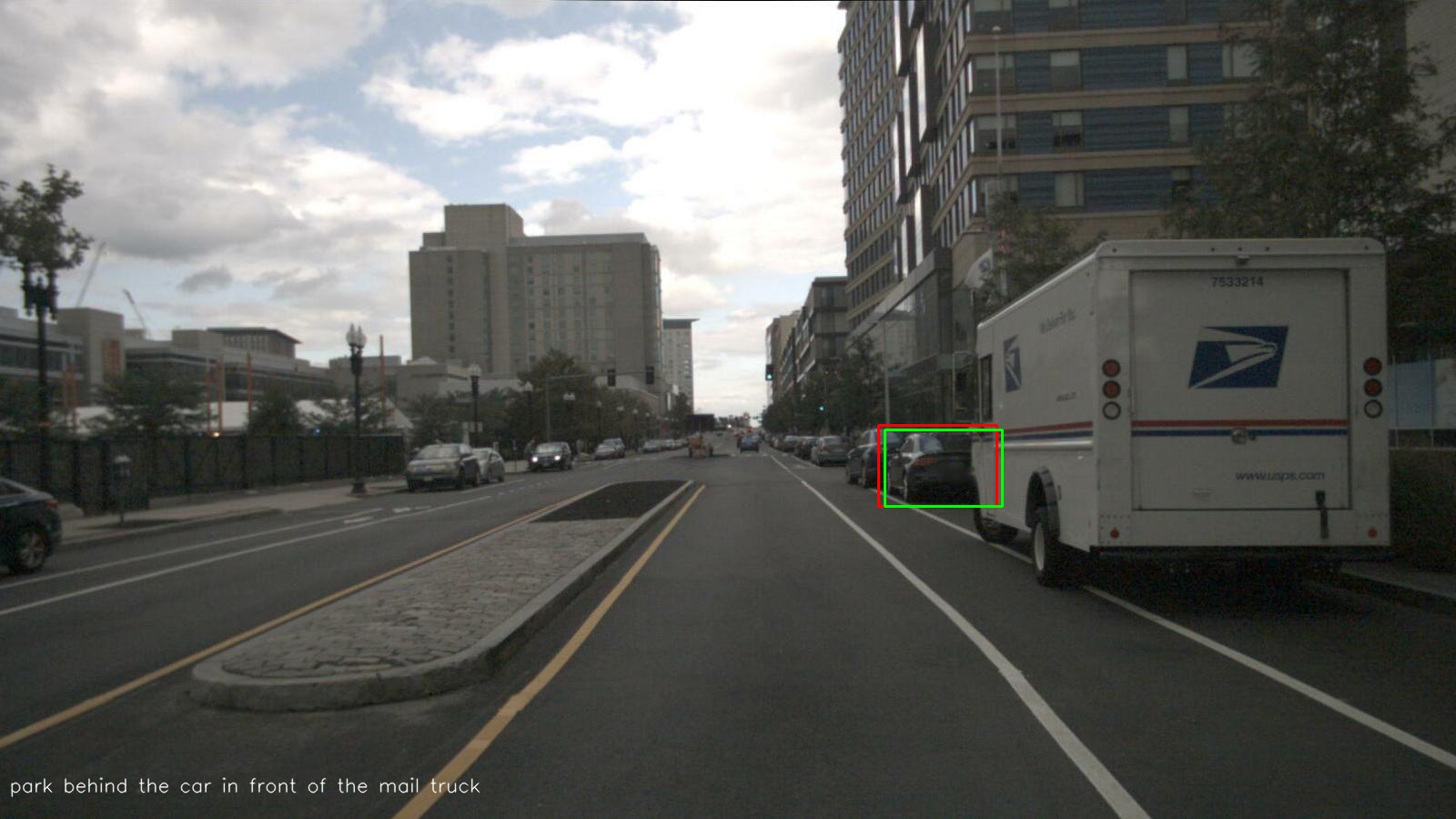}}
\end{minipage}

\begin{minipage}{0.3\linewidth}
\vspace{3pt}
    ``i see my friend mark. stop here for a moment"
\end{minipage}
\begin{minipage}{0.3\linewidth}
\vspace{3pt}
    ``pull up next to the guy wearing the safty vest so i can talk to him"
\end{minipage}
\begin{minipage}{0.3\linewidth}
\vspace{3pt}
    ``park behind the car infront of the mail truck"
\end{minipage}

\caption{\textbf{Visualization results on Talk2Car.} The green bounding box in the image corresponds to the ground truth, while the red bounding box represents the output generated by our model. The instruction corresponding to the image is provided below.}
\label{fig:qualitative}
\end{figure*}

The qualitative results of our model on Talk2Car are shown in Fig.~\ref{fig:qualitative}. From the image, it is evident that our model demonstrates a certain level of reasoning ability. It accurately comprehends the target and successfully detects the object.

\section*{Future work}
For now, we only use Bert as our language model considering inference and deployment efficiency. In the future, we would try to replace a LLM instead of Bert. 

\bibliography{references}
\bibliographystyle{plain}

\end{document}